\documentclass[letterpaper, 10 pt, conference]{ieeeconf}  % Comment this line out if you need a4paper

\IEEEoverridecommandlockouts                              % This command is only needed if 
\usepackage[numbers,sort&compress]{natbib}
\usepackage{multirow}
\usepackage{graphicx}
\usepackage[colorlinks=true, linkcolor=blue, citecolor=blue, urlcolor=blue]{hyperref}
\usepackage{caption}
\usepackage{subcaption}
\usepackage{float}
\usepackage{threeparttable}
\usepackage{amsfonts}
\title{\LARGE \bf
Flow-Enabled Generalization to Human Demonstrations \\ in Few-Shot Imitation Learning
}
\author{
    Runze Tang$^{1}$ and Penny Sweetser$^{1}$
    \thanks{$^{1}$The Australian National University
    {\tt\small \{runze.tang, penny.kyburz\}@anu.edu.au}}
}

\begin{document}
% \maketitle
\thispagestyle{empty}
\pagestyle{empty}

\twocolumn[{%
\renewcommand\twocolumn[1][]{#1}%
\maketitle
\begin{center}

\includegraphics[width=\textwidth]{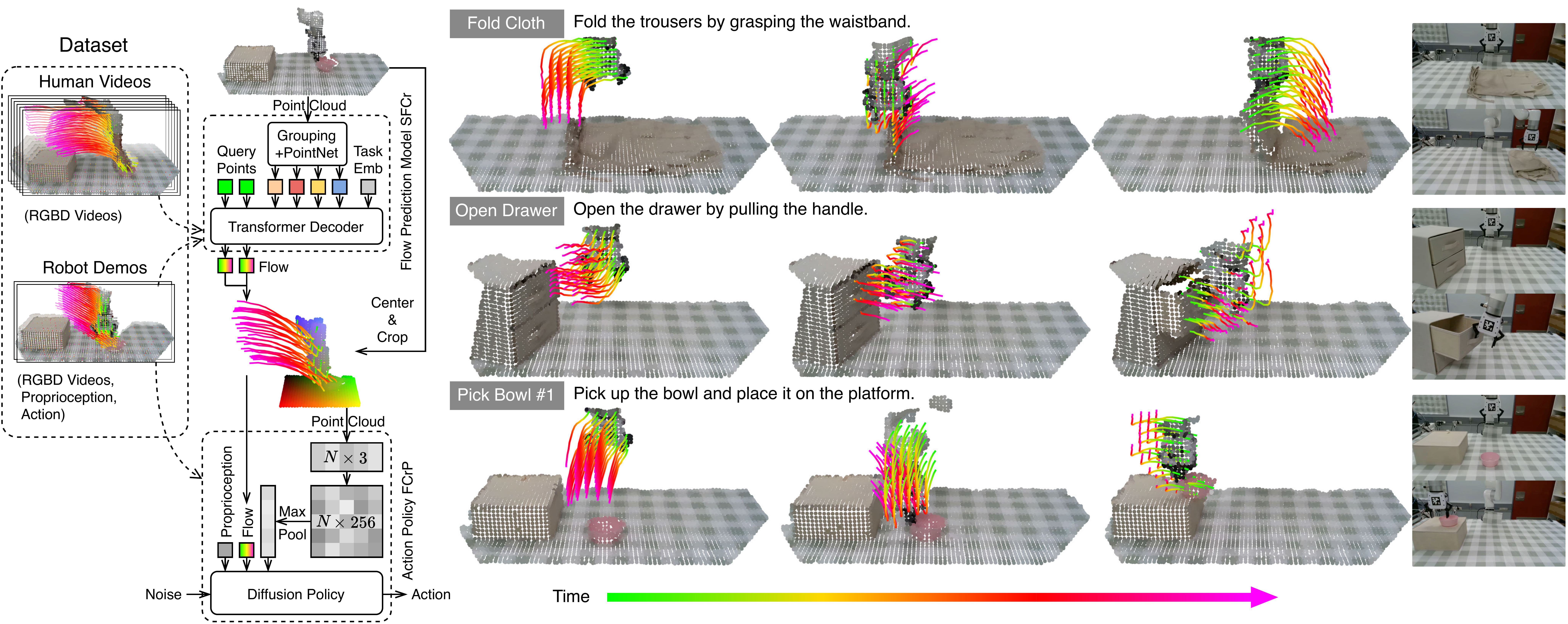}

\captionof{figure}{\textbf{Overview of our work.} Left: We use 30 human videos and 10 robot demonstrations to train the cross-embodiment flow prediction model SFCr and the flow-conditioned policy FCrP. Right: The point cloud observation from a single third-person-view camera and the predicted flow during execution. The images on the right are the beginning and success states of each task. }
\label{fig_main}
\end{center}
}]
\footnotetext[1]{The Australian National University {\tt\small \{runze.tang, penny.kyburz\}@anu.edu.au}}

%%%%%%%%%%%%%%%%%%%%%%%%%%%%%%%%%%%%%%%%%%%%%%%%%%%%%%%%%%%%%%%%%%%%%%%%%%%%%%%%
\begin{abstract}
Imitation Learning (IL) enables robots to learn complex skills from demonstrations without explicit task modeling, but it typically requires large amounts of demonstrations, creating significant collection costs. 
Prior work has investigated using flow as an intermediate representation to enable the use of human videos as a substitute, thereby reducing the amount of required robot demonstrations. However, most prior work has focused on the flow, either on the object or on specific points of the robot/hand, which cannot describe the motion of interaction. 
% both the robot and the objects. 
% Moreover, using flow as a guide to achieve generalization to scenarios seen only in human videos remains underexplored, as flow cannot sufficiently represent detailed action information. 
Meanwhile, relying on flow to achieve generalization to scenarios observed only in human videos remains limited, as flow alone cannot capture precise motion details. 
Furthermore, conditioning on scene observation to produce precise actions may cause the flow-conditioned policy to overfit to training tasks and weaken the generalization indicated by the flow.
% Although adding scene observation as conditions provides sufficient information to predict accurate actions, it could lead to overfitting on training tasks.
% , thereby reducing the generalization ability.
% Our method addresses these gaps by balancing the reliance between flow and scene observation. 
To address these gaps, we propose SFCrP, which includes a \underline{S}cene \underline{F}low prediction model for \underline{Cr}oss-embodiment learning (SFCr) and a \underline{F}low and \underline{Cr}opped point cloud conditioned \underline{P}olicy (FCrP).
SFCr learns from both robot and human videos and predicts any point trajectories. FCrP follows the general flow motion and adjusts the action based on observations for precision tasks. 
% Our method outperforms SOTA baselines across three real-world tasks, while also exhibiting strong spatial and instance generalization to scenarios seen only in human videos. 
Our method outperforms SOTA baselines across various real-world task settings, while also exhibiting strong spatial and instance generalization to scenarios seen only in human videos.

\end{abstract}

%%%%%%%%%%%%%%%%%%%%%%%%%%%%%%%%%%%%%%%%%%%%%%%%%%%%%%%%%%%%%%%%%%%%%%%%%%%%%%%%
\section{INTRODUCTION}
As a fundamental subclass of Imitation Learning (IL), behavior cloning (BC) is a widely adopted strategy for offline policy learning from demonstrations \citep{zare2024ILsurvey}. 
While BC can address tasks with complex dynamics, 
% such as deformable object manipulation, 
it usually requires dozens
% to hundreds 
of demonstrations for simple tasks \citep{wang2024rise, chi2023diffusionpolicy, chen2024SUGAR}, and
% hundreds of 
thousands to achieve robust generalization
% across diverse environments 
\citep{brohan2022rt1, zitkovich2023rt2}.
% To achieve robust generalization across diverse environments, the scale of demonstrations needed can increase to hundreds of thousands \citep{brohan2022rt1, zitkovich2023rt2}. 
However, collecting large-scale datasets
% , even via crowdsourcing, 
is often cost-prohibitive given the specialized equipment for each demonstrator \citep{brohan2022rt1, zitkovich2023rt2, hagenow2024versatile}. 
Therefore, many studies focus on using videos of human manipulation as a substitute for robot demonstrations \citep{bharadhwaj2024towards, ren2025MTpi, yuan2024gflow}.

% \vspace{-0.01mm}
With 3D spatial information, point cloud-based methods generally have better data efficiency and generalization capabilities than image-based methods \citep{zhu2024pcmatters}. 
% Images provide pixel-level precision in capturing object boundaries and inter-object relationships, but lack depth information. Point clouds provide 3D spatial information but tend to be noisy at object edges due to hardware limitations.
% Furthermore, 
% Due to the nature of image convolution, 
The image convolution features of the robot and human hand regions are inevitably different. Therefore, distribution adaptation methods are required to align image representations \citep{wang2023mimicplay, ren2025MTpi}. While in a point cloud, the robot or human hand is represented by points in the air.
% , which are rarely processed jointly with the surrounding context. 
Combined with segmentation methods, point clouds become a well-suited representation for cross-embodiment learning \citep{yuan2024gflow}. 
Flow, that is, the trajectories of points, has been confirmed to be an effective representation that describes motions to bridge human
% video
and robot demonstrations \citep{ren2025MTpi, yuan2024gflow, xu2024i2f2a, wang2025SKIL, zhi20253dflowaction}. 
However, most prior work has focused on flow on the object or robot arm solely to achieve cross-embodiment learning. Focusing solely on the object's motion often overlooks the robot's pre-grasp motion \citep{yuan2024gflow}. Conversely, only considering the motion of the robot omits the details of the interaction with the object \citep{ren2025MTpi}. In this work, we present a point cloud-based method that enables any-point flow prediction and cross-embodiment learning.

For point cloud scene perception, using a Transformer model to process the point cloud represented by tokens could effectively capture the spatial information of objects in the scene \citep{chen2024SUGAR, wang2024rise}, but lacks point-level details. Point-level perception methods, such as DP3 \citep{ze2024DP3}, often struggle in identifying scene changes, resulting in limited generalization \citep{ke2024Act3d}. Moreover, diffusion policy is shown to tend to overfit the training tasks and lacks generalization ability \citep{he2025demystifying, wu2025afforddp}. 
To address these issues, we use flow as an intermediate representation to bridge a Transformer-based flow prediction model and a flow-conditioned action policy with point-level perception. 
We crop the point cloud observation of the policy and balance the reliance between the point cloud and the flow to reduce overfitting. 
% We empirically demonstrate that using flow as a condition could significantly enhance the diffusion policy's generalization ability.
% by redirecting the reliance from point cloud to flow.  
% We analyze the contributions of three key components of our policy network through three research questions: 
% We empirically demonstrate that our method can address the aforementioned issues. 
% We summarize our main contributions as:
In summary, our main contributions are: 
\begin{itemize}
    \item A flow prediction model SFCr that predicts any point trajectories with high cross-embodiment data efficiency.    
    \item A flow and cropped point cloud conditioned policy FCrP that achieves spatial and instance generalization. 
%     % \item A \textbf{F}low and \textbf{L}ocal cropped point cloud conditioned \textbf{P}olicy (FLP) that could achieve both spatial and instance generalization. 
    \item Comprehensive experiments that demonstrate flow is a representation that can (1) bridge group-level spatial relationship perception and point-level detail recognition, (2) align robot demonstrations and human videos, and (3) significantly reduce overfitting of diffusion policy to achieve stable generalization. 
\end{itemize}

% We analyze how our methods address the gaps through three research questions:
% Furthermore, we analyze our method through four research questions:
% RQ1: How effective is segmentation in narrowing the cross-embodiment appearance gap?
Furthermore, we discuss our method, including the underlying mechanisms, and the aforementioned issues in the broader field through four research questions:
RQ1: How effectively does segmentation narrow the cross-embodiment appearance gap?
RQ2: Through what mechanism does cropping the point cloud enhance the policy performance in precision tasks? RQ3: How does conditioning on flow enhance policy generalization? RQ4: How effective is balancing the reliance in alleviating overfitting of the diffusion policy?
% research questions focusing respectively on the flow, the local point cloud observation, and the balance of reliance on them. 
% Together, they provide a comprehensive evaluation of the effectiveness.

\section{Related Work}

% big table - RGB/PC, pretrain? human video embedding? obj flow? eef flow? CV detection? tracking? segmentation? generalization?
Behavior cloning in imitation learning is a framework where agents acquire skills by observing and replicating expert behaviors \citep{bain1995BC, atkeson1997ILrobot}. 
In behavior cloning for robot manipulation, RGB images \citep{chi2023diffusionpolicy, zhao2023ACT, goyal2023rvt, gervet2023act3d, huang2024a3vlm} and point clouds \citep{ze2024DP3, wang2024rise, chen2024SUGAR, chen2023polarnet, shridhar2023peract, gervet2023act3d, rashid2023language} are two typical types of observations. 
Some previous work has combined RGB image features and point clouds to obtain both pixel-level perception and depth information \citep{ke2024Act3d, wang2024GenDP, wang2023sparsedff, zhang2023SGR, zhang2024SGRv2}, but it required multiple cameras to mitigate the impact of convolution background blending and view-dependent features.
% blending background pixel features. 
% Works that predict key-points or trajectories on multi-view images require motion alignment to prevent failure caused by motion inconsistency among views \citep{ren2025MTpi, goyal2023rvt}.
% particularly between closely located objects
In this work, we focus on the point cloud obtained from a single third-person view camera.

% convolution/group
% Convolution is a widely used method to obtain image features, which preserves both local texture and global spatial information. For point clouds, the similar approaches are spatial convolution \citep{SpatialConv} and sampling and grouping (SG) \citep{guo2021pct}. 

To enhance performance, many papers went beyond using manipulation task data. Some leveraged large-scale image or point cloud datasets for pretraining \citep{brohan2022rt1, majumdar2023vc1, chen2024SUGAR, radosavovic2023mvp, jing2023exploring, qian20243dmvp, zhang2023SGR}, while others directly used pre-trained visual models as feature extractors \citep{oquab2023dinov2, wang2025SKIL, wang2024GenDP, zitkovich2023rt2, shafiullah2022clipfields, huang2024a3vlm, rashid2023language, wang2023sparsedff}. Alternatively, some papers relied on in-domain manipulation task data but trained on more diverse forms, such as data collected from different robot embodiments \citep{zhang2024VTK, brohan2022rt1, dasari2019robonet, jiang2024rpr} or human videos \citep{mccarthy2025towards} through visual embeddings \citep{nair2022r3m, yang2024spatiotemporal, radosavovic2023mvp, bharadhwaj2024towards, jing2023exploring}, object flow \citep{zhi20253dflowaction, wang2025SKIL, wu2021vat, yuan2024gflow, xu2024i2f2a}, or human hand trajectory \citep{wang2023mimicplay, ren2025MTpi, bharadhwaj2023zero, shaw2024learning, qiu2025humanoid}. 
% \citep{ye2024latent}. 

Existing approaches to predict flow can be classified into three main categories: predicting flow on the object \citep{eisner2022flowbot3d, zhi20253dflowaction, wang2025SKIL,seita2023toolflownet,wu2021vat,yuan2024gflow,xu2024i2f2a}, predicting the flow on the robot arm \citep{wang2023mimicplay, ren2025MTpi, zhang2024VTK, chen2025ec}, and predicting trajectories of any points in the scene \citep{wen2023ATM, bharadhwaj2024track2act, gao2024flip}. 
% Considering only the flow of points on the object and ignoring the robot is a good way to bridge robot demonstrations and human videos. 
% Under the flow of object setting, \citet{seita2023toolflownet} proposed a consistency loss to ensure that trajectories of points on the object follow the same SE(3) transformation. 
% However, it lacks information about robot motion and may struggle with deformable objects. 
Considering only the flow of the object could easily achieve cross-embodiment learning, but it lacks information about the robot motion, especially before grasping \citep{yuan2024gflow}.
While the trajectory of points only on the robot cannot capture the interaction with objects. 
% Approaches such as Any-point Trajectory Modeling (ATM) \citep{wen2023ATM} 
Although predicting the flow of the entire scene could capture any motion in the scene, it has a larger cross-embodiment gap \citep{xu2024i2f2a}. 
Prior work has shown that using large-scale video datasets with diverse embodiments could address the cross-embodiment gap in predicting any-point trajectories \citep{bharadhwaj2024track2act}. In this paper, we address this gap with a small-scale dataset using a well-designed flow prediction method. 
% However, the task difficulty also increases accordingly. 
% Our method aims to predict any-point trajectory in the scene, while achieving cross-embodiment generalization.
% with a small-scale dataset.
% utilizing both human and robot demonstrations.
% But our method could also utilise human videos to guide both robot motion and interaction with the object. 

The choice of observation can be critical for the action policy \citep{mandlekar2021robomimic}. 
% To obtain actions from the predicted flow, 
Relying solely on the predicted flow as observation, some researchers compute actions in a heuristic way \citep{eisner2022flowbot3d, zhi20253dflowaction, seita2023toolflownet, wu2021vat, yuan2024gflow, ren2025MTpi}. In contrast, other methods used a policy network to produce actions conditioned on the predicted flow and scene observation \citep{xu2024i2f2a, wang2023mimicplay, wen2023ATM, bharadhwaj2024track2act}. Policies that use both flow and scene observation as conditions are capable of performing more precise actions and correcting potential inaccuracies in the input flow. In comparison, heuristic methods highly rely on the accuracy of the predicted flow. However, the scene observation can undermine the generalization ability of the action policy. In contrast, many flow-conditioned heuristic action policies that are not conditioned on the scene observation show a stronger generalization ability \citep{eisner2022flowbot3d, zhi20253dflowaction, wu2021vat, yuan2024gflow, ren2025MTpi}.

Most of the flow-conditioned policy models are actually conditioned on the flow feature vector rather than the raw flow \citep{xu2024i2f2a, wang2023mimicplay, wen2023ATM}. 
% In their method, the flow is more like a 
They use flow as a training target of the flow prediction model, rather than a representation that truly bridges the flow prediction and the action policy. 
This could further mitigate the failure caused by inaccurate flow, but could lead to potential overfitting to the training task and a strong binding between the flow model and policy model. 
Our approach uses the raw flow and a local cropped point cloud as the condition for the action policy. In this way, we ensure heuristic-level generalization based on the flow and obtain enough observations for precision-demanding tasks.

% Most existing approaches that utilize flow for cross-embodiment employ off-the-shelf models for hand detection \citep{ren2025MTpi, wang2023mimicplay, bharadhwaj2023zero, shaw2024learning}, segmentation \citep{wang2025SKIL, yuan2024gflow, zhi20253dflowaction, xu2024i2f2a, bharadhwaj2024towards}, and/or trajectory tracking \citep{wang2025SKIL, yuan2024gflow, xu2024i2f2a}. We use CoTracker \citep{karaev23cotracker} to obtain point trajectories and FastSam \cite{FastSam} to segment robot/human hands, evaluating how effectively segmentation narrows the gap between robot and human appearance. 
% The cross-embodiment based on image observation usually uses distribution loss to force the embedding space of the robot and hand to overlap \citep{ren2025MTpi, wang2023mimicplay}. Due to the nature of convolution operations on images, the robot arm or human hand regions are inevitably blended with surrounding content. While in a point cloud, the robot arm or human hand are represented by points in the air. In most cases, operations such as spatial convolution \citep{SpatialConv} or point cloud grouping \citep{guo2021pct, qian2022pointnext, chen2024SUGAR} using farthest point sampling (FPS) and K-nearest neighbour (KNN) will not give a local feature that contains both the robot/hand and the surrounding contents. Therefore, we believe point clouds could be a better representation for cross-embodiment learning than images. 
% In our work, we use point cloud grouping together with robot/hand segmentation to narrow the gap between robot and human video. 

\section{Approach}
Our method consists of two parts: the flow prediction model SFCr and the flow-conditioned policy FCrP. SFCr can learn from human videos, while FCrP is trained solely on robot demonstrations. 
% The architecture of the networks is shown in Fig.~\ref{fig_main}. 
% In the following sections, we introduce the data preprocessing and the model details.
% The flow prediction network takes the point cloud observation of the whole scene and $n$ query points as input to predict the future trajectory of the query points. 
% \subsection{Dataset Preparation}
Each robot demonstration includes an RGBD video, the corresponding robot proprioception, and the associated actions. We use the position of the gripper and its two fingers as proprioception data $g\in \mathbb{R}^{3\times3}$. The human demonstrations contain only the RGBD videos. 
We first build the raw point cloud of each RGBD image, then use voxel downsampling to reduce the number of points. 
To obtain the ground truth flow, we use CoTracker \citep{karaev23cotracker} to track grid-sampled query points in the RGB video and map them to the raw point cloud to obtain 3D point trajectories $F_{0:T}\in\mathbb{R}^{T\times 3}$ to form the flow $\mathcal{F}$ of the whole scene. To segment the robot and human hand, we apply FastSam \citep{FastSam} to
each frame of the RGB videos. 
% We segment the human video using a language prompt and robot videos using the bounding boxes computed from the AprilTag on the robot.
We segment human videos using a language prompt at the first frame and bounding boxes for the rest. 
The robot video is segmented using bounding boxes based on the April tag on the robot gripper as prompts. 
% For the first frame of each human video, we obtain the hand segmentation using language instructions. For the following frames, we use the bounding box of the hand segmentation in the previous frame as a prompt. 
% For robot videos, we obtain bounding boxes based on the AprilTag on the robot as prompts. 

\subsection{SFCr: Cross-Embodiment Scene Flow Prediction Model}

% The flow is the 3D point trajectory of the points in the scene. We follow the same assumption as ATM \citep{wen2023ATM} to predict any points in the scene. Given the point cloud observation of the whole scene and the query points, the network predicts the future trajectories of the query points. The observation horizon is 1. The flow prediction model aims to utilise both robot demonstrations and human videos to predict a rough flow of the scene. The accurate actions are based on the action policy that will be introduced in the next section. 

The general architecture of our flow prediction network is shown in Fig.~\ref{fig_main}.
% We use a Transformer decoder that incorporates both self-attention and cross-attention as its main component. 
We use a Transformer decoder \citep{vaswani2017attention} as its main component. 
The input tokens include point cloud tokens, task embedding, and flow query tokens. Each point cloud token is the PointNet \citep{PointNet} features of a local group of points, adding the spatial encoding. 
% We follow the sampling and grouping method \citep{guo2021pct, chen2024SUGAR} to 
We sample the center $x_0$ of each group by farthest point sampling and select points near the center to form the group $x_{0:k}$. 
% A shared PointNet is used to calculate the local feature of the local points centered at their center. And a shared MLP is applied to the position of the group centers to obtain the spatial encodings. 
Each flow query token is the spatial encoding of the starting point $F_0$ of the corresponding trajectory $F_{0:T}$. 
% We use the same MLP to encode the point cloud group center position and the flow query points position. 
The point cloud group center $x_0$ and the flow query points $F_0$ share the same spatial encoder. 
The output of each flow query token is fed into a shared multi-layer perceptron (MLP) to obtain the predicted trajectory. By using the Transformer decoder to process all these inputs, we expect the model to match the task embedding and the query points with the point cloud group tokens, learn a rough motion for each group, align across flow tokens, and finally provide a refined query point motion represented by trajectories.

% The flow representation can highly affect the prediction accuracy. 
We evaluate the predicted flow based on the absolute position. 
But for each trajectory point, we use the position related to its query point as the predicting target $F_{i}-F_0$ and minimize the L1-norm loss.
% We found that t
Empirically, this representation has a lower prediction error than position related to the previous point $F_{i}-F_{i-1}$ \citep{yuan2024gflow} or absolute position $F_{i}$. 
% We compare absolute position (ABS), position related to the query point (starting point) (RE0), and position related to the previous positions (REE) in Appendix.  

% flow num
% To reduce the Transformer calculation complexity and memory usage, 
To prevent the model from overfitting to the spatial distribution of query points, our flow prediction model is trained on a trajectory subset of $N_q=64$ query points for each point cloud observation. 
% However, among all the points in the scene, 
However, more than half of the points in the scene remain static. 
Randomly sampling from all trajectories will result in an imbalanced distribution of trajectory lengths. 
% flow select
% Since our flow prediction model predicts the trajectory of any points in the scene, and more than half of the points in the scene remain static, we cannot ignore the imbalance of the trajectory length. 
Therefore, for each sample, we first sample a moving ratio $p_m \sim \mathcal{U}(0, 1)$, then select $ p_mN_q$ points whose future trajectory is not static and $(1-p_m)N_q$ static points as query points. 
% We determine whether a query point is static or not by comparing the width of a trajectory with a hyperparameter. The width of a trajectory $F$ is defined as $\max_{i,j\in \left[0,T\right)} \|F_i-F_j\|_2$, the largest distance between any two points $F_i$, $F_j$ on the trajectory $F$. 
We determine whether a query point is static or not based on the width of a trajectory, which is defined as $\max_{i,j\in \left[0, T\right)} \|F_i-F_j\|_2$, the largest distance between any two points $F_i$, $F_j$ on trajectory $F$. This metric is more noise-robust than the accumulated displacement. 
During execution, the query points are selected via grid-based sampling from the box-shaped cropped point cloud centered at the robot gripper. 
We feed all the in-box query points in at once to maintain consistency among trajectories. 
To minimize the visual difference between the robot and the hand and enable cross-embodiment flow prediction, we segment the image to obtain robot/hand segmentation. We replace the point cloud color in the robot/hand region with (1,0,1) and add a dimension after the XYZRGB values to indicate whether a point belongs to the robot/hand or not. Moreover, we randomly remove a fraction of point cloud group tokens where most points are marked as robot/hand. This aims to train the flow prediction model not to remember the exact shape of the robot/hand, but to make inferences based on their approximate position. 

% \vspace{-2pt}
\subsection{FCrP: Flow and Croped Point Cloud Conditioned Policy}

Our diffusion-based \citep{chi2023diffusionpolicy, ho2020diffusionmodel} action policy produces actions by progressive denoising, conditioned \citep{perez2018film} on the predicted flow  $\mathcal{F}$ and state observations $\{s_f,s_{t-1},s_{t}\}$, as Fig.~\ref{fig_main} shows. The observation horizon includes three states: (1) the flow state $s_f$, at which the flow $\mathcal{F}$ is predicted, (2) the state before the current state $s_{t-1}$, and (3) the current state $s_t$, where $t\geq f$. Each state observation includes the local cropped point cloud $X$ and proprioception data $g$. 
% Our action policy is built upon DP3 \citep{ze2024DP3}. 
We use the DP3 \citep{ze2024DP3} encoder as the point cloud perception model, which first calculates the features of each point using shared MLPs, then applies max pooling to the feature dimension, followed by another MLP layer to obtain a compact representation. 
% The decision part of DP3 is a diffusion policy that denoises the action with the point cloud feature and the robot state feature as conditions. The robot state is the position of three points, one on the robot arm gripper and two on the fingers of the gripper. 

% We use a shared MLP to calculate the flow feature as an additional condition for the diffusion policy. 
Instead of using the point cloud observation of the whole scene, we crop the point cloud observation $X$ to keep only a box-shaped region around the robot gripper and center it with the robot gripper as the origin for each state $\{s_f,s_{t-1},s_{t}\}$. For the proprioception points $\{g_f,g_{t-1},g_t\}$ at each state and the flow $\mathcal{F}$ at the flow state $s_f$, we center them around the gripper position at the flow state $s_f$, to keep the related spatial information within the observation horizon. In this way, the observation of our action policy is fully localized with the robot gripper as origin without any absolute spatial information, which enables the generalization following the flow and action adjustment conditioned on the local point cloud. In execution, we select query points within the box-shaped cropped point cloud for flow prediction.
% and only keep the non-static ones as the flow condition. 

% align
% \textbf{Flow aligning.}
To decouple our action policy from the flow prediction network in terms of inference frequency, we introduced a flow-state-action alignment mechanism. Our action policy predicts a sequence of actions starting from the flow state $s_f$ with an execution mask that matches the motion between the flow state $s_f$ and the current state $s_t$ with the flow $\mathcal{F}$. 
% Intuitively,
The execution mask indicates which actions were performed from the flow state $s_f$ to reach the current state $s_t$, and which actions should be performed next. This enables the prediction of an arbitrary number of actions from the same predicted flow.  Predicting actions starting from the flow state temporally aligns the motion in flow and the actions, enabling our action policy to produce actions that follow the same motion as the flow to enable generalization. 
% learn the similarity between the actions and the flow. 
% In training, after selecting the flow step $s_f$, we randomly select the current step $s_t$ within the next 32 steps of the flow step $s_f$. 

% align exec
Apart from enabling parallel inference and asynchronous flow condition updating by conditioning on the previous flow and replacing it whenever the new one is predicted, our flow-state-action alignment mechanism could also enable heuristic flow prediction skipping at error-prone states. For example, when the gripper is close to the object before grasping, if the flow prediction model mistakenly thinks that the robot has already grasped the object and gives a flow with an upward direction, the action policy may fail to grasp the object. 
% The primary benefit of decoupling the inference frequency between the flow prediction model and the action policy is to mitigate the impact of inaccurately predicted flows. 
% The flow prediction model has only the point cloud observation at flow state $s_f$.
% without proprioception information $g$. 
% In addition to the group-removal augmentation, it may not be clear whether the gripper is closed or the object is successfully picked up. 
% When the gripper is close to the object to pick up, if the flow prediction model mistakenly thinks the robot has grasped the object and gives a flow going in another direction, the action policy may fail to grasp the object. 
% With the help of the flow-alignment mechanism that enables an arbitrary number of actions to be predicted based on the same flow, 
% Therefore, we skip the flow prediction if the gripper closing actions are predicted in the near future steps. 
% With the flow-alignment mechanism enabling the prediction of an arbitrary number of actions from the same flow, we omit flow prediction if the upcoming actions involve gripper closure.
% In this way, we prevent the model from re-predicting the flow when approaching the grasping action, to reduce failures caused by incorrectly predicted flows. 

% \textbf{Train with predicted flow.}
Many prior approaches use the predicted flow embeddings as the condition of the policy network. In this way, the policy network becomes more robust to the inaccurately predicted flow. To achieve a similar goal using the raw flow, we let a trained flow prediction model predict the flow for each robot demonstration, to replace the ground-truth flow during action policy training. Empirically, we do not observe a significant performance drop using different flow prediction models for predicted flow generation and execution. 
% The resulting policy network will be more robust in the input flow. 
% aug PC0
Moreover, we balance the condition reliance between the point cloud and the flow to reduce overfitting caused by point cloud conditioning. We randomly mask the point cloud (MP) by replacing the entire point cloud with zero with a probability of 0.5. Thus, the policy is forced to rely more on the flow.
% , which strengthens the policy's generalization ability by following the flow. 

\begin{table}
\caption{Task Properties}\label{tab_task}
\begin{tabular}{lccc}
\hline
Key Property & Fold Cloth & Open Drawer & Pick Bowl \\ \hline
Multi-Task & No & No & Yes \\
Object Type & Deformable & Articulated  & Rigid \\
Include Grasp Action & Yes & No & Yes \\
Action Accuracy Req. & Medium & High & Low \\ \hline
\end{tabular}

\end{table}

% \begin{table}[]
% \caption{Pick Bowl Task Setting}\label{tab_task_pick}
% \begin{tabular}{lllllllll}
% \hline
% \multirow{2}{*}{Setting} & \multicolumn{8}{c}{Pick Bowl} \\ \cline{2-9} 
% & \#0 & \#1 & \#2 & \#3 & \#4 & \#5 & \#6 & \#7 \\ \hline
% Bowl Type & Pink & Pink & Pink & Pink & Pink & Pink & Pink & Yellow \\
% Bowl Position & 1 & 2 & 3 & 4 & 5 & 6 & 7 & 7 \\
% \#Robot Demo & 10 & 10 & 10 & 10 & 10 & 0 & 0 & 0 \\
% \#Hand Demo & 30 & 30 & 30 & 30 & 30 & 30 & 30 & 30 \\ \hline
% \end{tabular}
% \end{table}
\begin{figure}

    \centering

    \includegraphics[width=\linewidth]{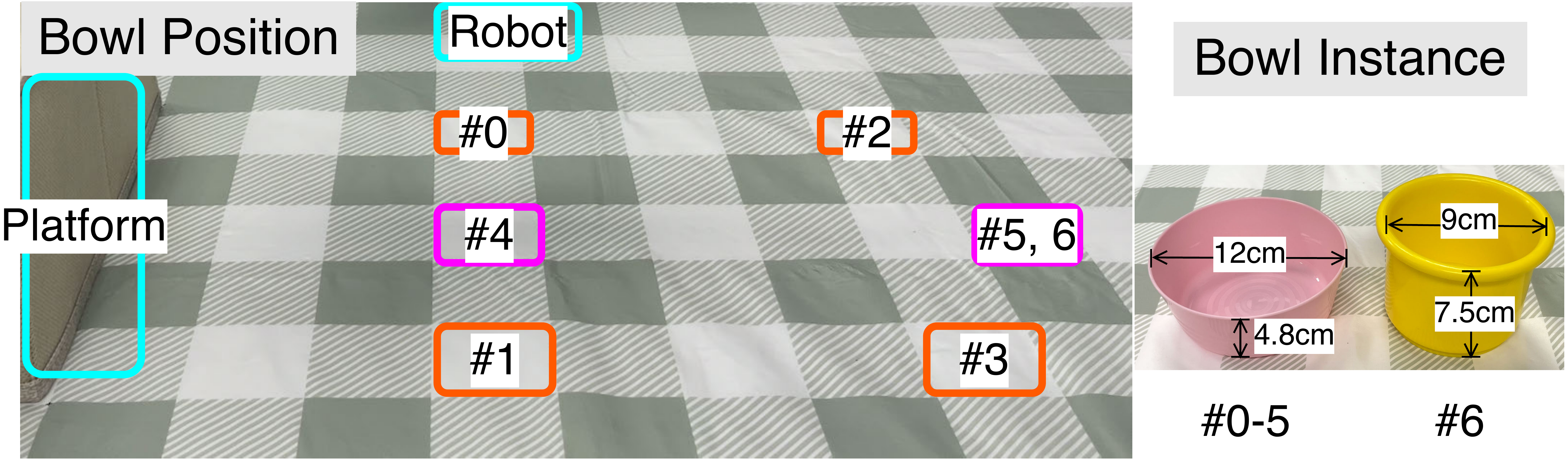}
    \caption{The bowl position (warm color rectangles) and instances of Pick Bowl tasks. 
    There are no robot demonstrations for \#4-6. 
    % The yellow bowl is taller with a smaller rim diameter than the pink bowl. 
    }
    \label{fig_task_pick}
\end{figure}

\section{Real-World Evaluation}
Fig.~\ref{fig_main} and Table~\ref{tab_task} list the settings and description of the real-world tasks. 
% We evaluate our method on three real-world tasks, whose setting is listed in Table~\ref{tab_task}.
% with different types of manipulating objects and action precision requirements (Table~\ref{tab_task}). 
Fig.~\ref{fig_task_pick} shows seven versions \#0-6 of the Pick Bowl task.
% The Pick Bowl tasks include seven versions, as Fig.~\ref{fig_task_pick} shows. 
% of which the settings are shown in Fig.~\ref{fig_task_pick}. 
For each task, we collect 10 robot demonstrations (R10) and 30 human videos (H30) for training, except Pick Bowl \#4-6, which have 30 human videos only.
% and no robot demonstrations. 
% Pick Bowl \#4 has a similar initial action direction as \#1, while that of \#5,6 is different from the others. 

% baselines
We compare the flow prediction accuracy of ours with ScaleFlow-L \citep{yuan2024gflow}, which is a 3D trajectory prediction model based on PointNeXt \citep{qian2022pointnext} and VAE \citep{higgins2017betavae, kingma2013vae}. 
We compare the manipulation task success rate of our method with DP3 \citep{ze2024DP3}, RISE \citep{wang2024rise}, and SUGAR \citep{chen2024SUGAR}. DP3 uses a well-designed point cloud encoder followed by a diffusion policy \citep{chi2023diffusionpolicy} to predict actions. RISE uses spatial convolution \citep{choy20194d} followed by a Transformer \citep{vaswani2017attention} to extract point cloud features for the final diffusion policy. SUGAR samples and groups the point cloud, extracting the group features as tokens for Transformer calculations. The decoder of SUGAR was designed for keypose action prediction. We adopted the action head of ACT \citep{zhao2023ACT} that is also based on the Transformer decoder, to enable SUGAR for micro-step action prediction. SUGAR pretrains the Transformer encoder on five pretraining tasks with four datasets ($\sim$940K samples). We evaluate SUGAR pre-trained using the ensemble of four datasets and from scratch. 
We also assess the ablated variants of our method, where the flow prediction network does not have robot/hand segmentation (w/o SG), the action policy has no point cloud observation (w/o PC), is not trained on the predicted flows (w/o PF), and does not mask the point cloud (w/o MP).

\begin{figure}
    \centering
    \includegraphics[width=0.97\linewidth]{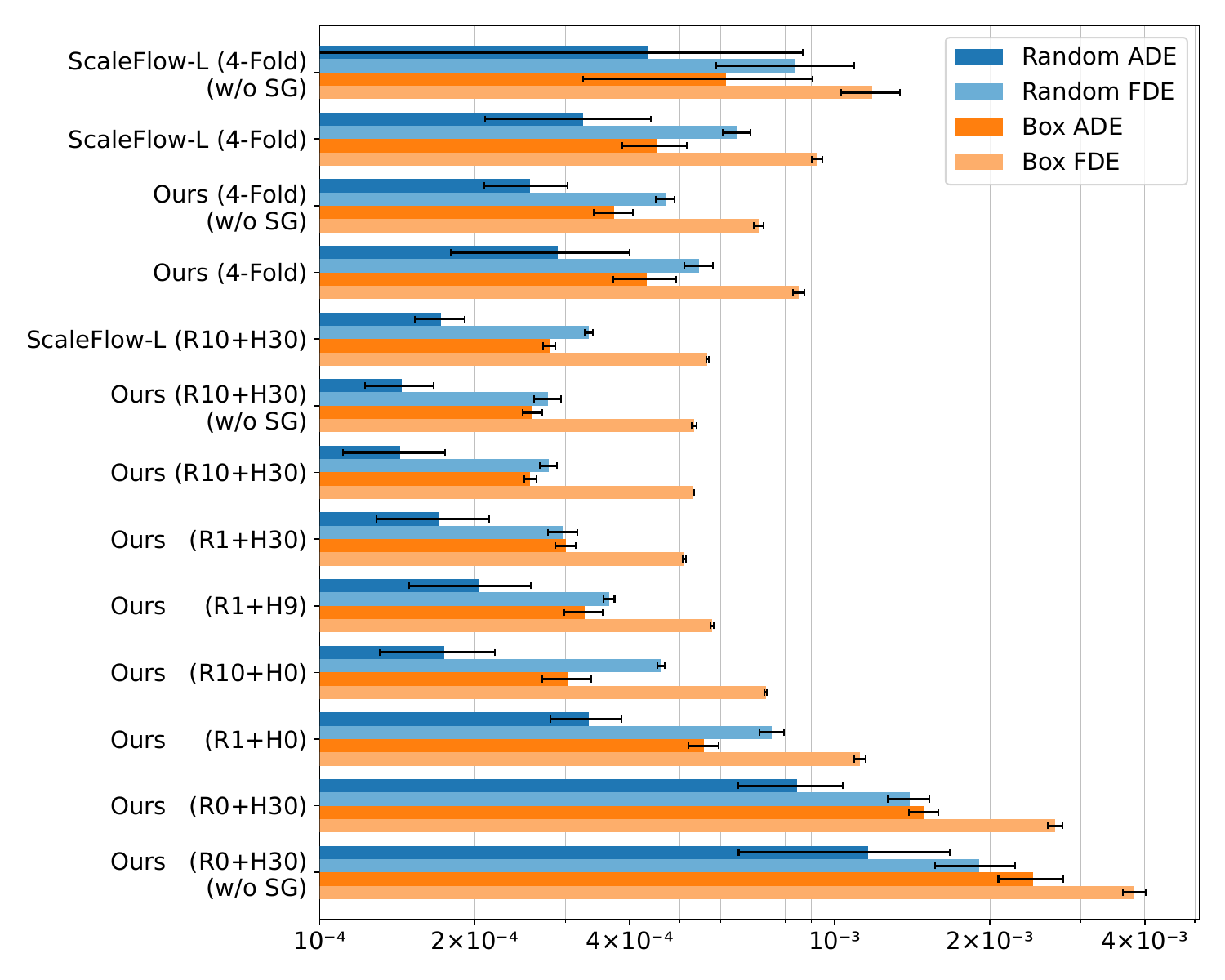}
    \caption{Logarithmic scale flow ADE and FDE over five seeds.}
    \label{fig_flow_ade_fde}
\end{figure}

\begin{table*}
\caption{\textbf{Success rate with full dataset (R10+H30).} The seen average success rate does not include Pick Bowl \#4-6. We conduct 20 trials for each task with different seeds, except for tasks that are almost impossible to complete, which have 10 trials. The figures in parentheses are the number of failures in which the first stage was completed but failed the final stage. }\label{tab_succ_main}
\resizebox{\textwidth}{!}{
\begin{threeparttable}
\begin{tabular}{lclllllllll}
\hline
\multicolumn{1}{l}{\multirow{2}{*}{Method}} & \multirow{2}{*}{\begin{tabular}[c]{@{}c@{}}Seen\\ Succ \%.\end{tabular}} & \multicolumn{1}{c}{\multirow{2}{*}{\begin{tabular}[c]{@{}c@{}}Fold\\Cloth\end{tabular}}} & \multicolumn{1}{c}{\multirow{2}{*}{\begin{tabular}[c]{@{}c@{}}Open\\Drawer\end{tabular}}} & \multicolumn{7}{c}{Pick Bowl} \\ \cline{5-11} 
\multicolumn{1}{c}{} & & \multicolumn{1}{c}{} & \multicolumn{1}{c}{} & \#0 & \#1 & \#2 & \#3 & \#4 & \#5 & \#6 \\ \hline
DP3 \citep{ze2024DP3} & 74.17 & 11/20 & 15/20 & 12/20 (+5) & \textbf{20/20} & \textbf{20/20} & 11/20 & \hphantom{0}0/10 & \hphantom{0}0/10 & \hphantom{0}0/10 \\
RISE \citep{wang2024rise} & 50.00 & \hphantom{0}3/20 (+14) & \hphantom{0}9/20 (+2) & \hphantom{0}3/20 (+15) & 17/20 (+2) & 10/20 (+3) & 18/20 & \hphantom{0}0/10 & \hphantom{0}0/10 & \hphantom{0}0/10 \\
SUGAR \citep{chen2024SUGAR} & 75.83 & \hphantom{0}7/20 (+1) & \hphantom{0}8/20 (+1) & 19/20 & \textbf{20/20} & 19/20 & 18/20 & \textbf{19/20} (+1) & \hphantom{0}9/20 & \hphantom{0}0/10 \\
SUGAR (pre-trained) & 88.33 & \textbf{18/20} & \hphantom{0}9/20 & \textbf{20/20} & \textbf{20/20} & \textbf{20/20} & 19/20 & \textbf{19/20} (+1) & \hphantom{0}8/20 & \hphantom{0}0/10 \\
Ours (w/o PC) & 63.33 & \hphantom{0}2/20 (+15) & \hphantom{0}1/20 & \textbf{20/20} & 19/20 (+1) & 16/20 & 18/20 (+2) & \textbf{19/20} & 17/20 & 18/20 \\
Ours (w/o PF\&MP) & 90.00 & 13/20 (+7) & \textbf{19/20} (+1) & \textbf{20/20} & 19/20 & 18/20 & 19/20 & \textbf{19/20} & 17/20 & 13/20 (+1) \\
Ours & \textbf{96.67} & \textbf{18/20} (+1) & 18/20 (+1)\tnote{*} & \textbf{20/20} & \textbf{20/20} & \textbf{20/20} & \textbf{20/20} & \textbf{19/20} (+1) & \textbf{20/20} & \textbf{20/20} \\ \hline
\end{tabular}
\begin{tablenotes}
      \footnotesize
      \item[*] This task requires high action precision, which highly relies on the point cloud observation. We did not apply the MP augmentation. 
    \end{tablenotes}
\end{threeparttable}
}

\end{table*}

\begin{table}
\caption{\textbf{Success rate of Pick Bowl \#0-3 with limited demonstrations.} We conduct 10 trials for each task with different seeds. The figures in parentheses are the number of failures in which the task was completed halfway but failed at the final stage. } \label{tab_succ_few}
\begin{tabular}{llllllc}
\hline
\multirow{2}{*}{Method} & \multirow{2}{*}{\begin{tabular}[c]{@{}l@{}}\# Demo\\ per Task\end{tabular}} & \multicolumn{4}{c}{Pick Bowl} & \multicolumn{1}{l}{\multirow{2}{*}{Avg.}} \\ \cline{3-6}
 & & \#0 & \#1 & \#2 & \#3 & \multicolumn{1}{l}{} \\ \hline
DP3 \citep{ze2024DP3} & R1+H0 & \hphantom{0}1 (+3) & \hphantom{0}0 (+3) & \hphantom{0}3 & \hphantom{0}0 & 10\% \\
RISE \citep{wang2024rise} & R1+H0 & \hphantom{0}0 & \hphantom{0}0 & \hphantom{0}0 & \hphantom{0}0 & \hphantom{0}0\% \\
SUGAR \citep{chen2024SUGAR} & R1+H0 & \hphantom{0}7 (+2) & \hphantom{0}0 (+1) & \hphantom{0}6 & \hphantom{0}3 & 40\% \\
Ours (w/o MP) & R1+H0 & \hphantom{0}6 & \hphantom{0}6 (+1) & \hphantom{0}4 & \hphantom{0}2 & 45\% \\
Ours (w/o MP) & R1+H30 & \hphantom{0}7 & \hphantom{0}5 & \hphantom{0}6 & \hphantom{0}4 & 55\% \\
Ours & R1+H0 & \hphantom{0}\textbf{9} (+1) & \hphantom{0}2 (+5) & \hphantom{0}8 & \hphantom{0}\textbf{9} & 70\% \\
Ours & R1+H30 & \hphantom{0}7 (+2) & \hphantom{0}\textbf{6} (+1) & \textbf{10} & \hphantom{0}7 & \textbf{75\%} \\ \hline
\end{tabular}

\end{table}

\begin{table}
\caption{\textbf{First stage success rate of Open Drawer.} The first-try and retry success rate of the first stage of the Open Drawer task, that is, hooking the drawer handle. } \label{tab_succ_retry}
\begin{tabular}{lrrrr}
\hline
Method & \multicolumn{1}{c}{\begin{tabular}[c]{@{}c@{}}\# First-Try\\Success\end{tabular}} & \multicolumn{1}{c}{\begin{tabular}[c]{@{}c@{}}\# Retry\\Success\end{tabular}} & \multicolumn{1}{c}{\begin{tabular}[c]{@{}c@{}}First-try\\ Succ \%.\end{tabular}} & \multicolumn{1}{c}{\begin{tabular}[c]{@{}c@{}}Retry\\ Succ \%.\end{tabular}} \\ \hline
DP3 \citep{ze2024DP3} & 11 & 4 & 55.0 & 44.4 \\
RISE \citep{wang2024rise} & 10 & 1 & 50.0 & 10.0 \\
SUGAR \citep{chen2024SUGAR} & 9 & 0 & 45.0 & 0.0 \\
SUGAR (pre-trained) & 9 & 0 & 45.0 & 0.0 \\
Ours (w/o PC) & 0 & 1 & 0.0 & 5.0 \\
Ours (w/o PF\&MP) & 17 & 3 & \textbf{85.0} & \textbf{100.0} \\
Ours (w/o MP) & 17 & 2 & \textbf{85.0} & 66.7 \\ \hline
\end{tabular}
\end{table}

\subsection{Flow Prediction Evaluation}
Our predicted flow during execution shows correct motion information.
In the middle column of the flow plots shown in Fig.~\ref{fig_main}, we can observe that the trajectory of points on the object remains static when the robot is approaching (green to red) and then starts moving with the same motion as the robot (red to magenta). 
Fig.~\ref{fig_flow_ade_fde} shows the average displacement error (ADE) and final displacement error (FDE) of the predicted flow of randomly selected query points (Random) and in-box around the robot gripper (Box) on the test robot demonstrations. The 4-Fold validation involves sequentially excluding robot demonstrations from one of the four Pick Bowl tasks (\#0–3) in each fold, evaluating the model on the task that does not have robot demonstrations. Compared to ScaleFlow-L \cite{yuan2024gflow}, our method shows a lower error in both the full dataset (R10+H30) and the 4-Fold settings. Moreover, our method trained with R10+H0, R1+H9, and R1+H30 achieves similarly low errors, comparable to using the full dataset (R10+H30). These results underscore the high cross-embodiment data efficiency of our method.

\textbf{RQ1: How effectively does segmentation narrow the cross-embodiment appearance gap?}
% : How effectively does segmentation narrow the cross-embodiment appearance gap?
When no robot data is available (R0+H30), our method shows a substantial error drop compared to the version not using robot/hand segmentation (w/o SG). 
We note that the higher flow error of our method without robot data (R0+H30) is primarily due to human videos having a higher speed and longer flow, rather than incorrect motion in the predicted flow. 
However, the difference between with and without segmentation is not notable when robot data are available (R10+H30 and 4-Fold). 
We attribute Ours (4-Fold) having a slightly higher error than without segmentation (w/o SG) to the information loss caused by the segmentation-based point cloud removal augmentation. 
In summary, (1) our flow prediction model could generalize to scenarios only seen in human videos (4-Fold) without the need for segmentation, (2) with segmentation narrowing the appearance gap, our model could predict the flow of unseen embodiment with correct motion. 

% (2) segmentation plays a vital role in narrowing the appearance gap between the seen and completely unseen embodiment.
% (1) segmentation plays a vital role in narrowing the cross-embodiment appearance gap without robot data, (2) our method learns the similarity between human and robot appearances, generalizing to scenarios only seen in human videos (4-Fold) without the need for segmentation. 

% (1) segmentation plays a vital role in narrowing the gap between robot and human appearance when robot data is missing, 
% But when both robot and human data are available, the effect of segmentation is negligible. 
% When robot data is missing, segmentation plays an important role in narrowing the gap between robot and human appearance. 

\subsection{Real-World Robot Manipulation}
Table~\ref{tab_succ_main} lists the success rate of all tasks using the full dataset (R10+H30), except for Pick Bowl \#4-6 tasks (R0+H30). Our method achieves the highest success rate compared to baseline methods across all tasks, and demonstrates strong spatial and instance generalization to scenarios that are only seen in human videos (Pick Bowl \#4-6). 
% that do not use human videos. 

% few demo
Table~\ref{tab_succ_few} lists the success rate of Pick Bowl \#0-3 tasks with limited robot demonstrations. The corresponding flow prediction error is shown in Fig.~\ref{fig_flow_ade_fde}. 
Our method achieves a 70\% average success rate with only one robot demonstration per task.
% With only one robot demonstration for each task, our method achieves a 70\% average success rate. 
% For flow prediction, even a single demonstration contains a large number of trajectory samples that the model can learn from. 
For flow prediction, our training flow sampling method exploits the trajectories even in a single demonstration to increase the amount of training samples indirectly. 
For our policy network, the conditioned flow guides the approximate motion when the gripper is far from the target object. 
Moreover, the cropped point cloud when the robot gripper approaches the bowl in Pick Bowl \#0 and \#1 is very similar, which also holds for \#2 and \#3. 
The similarity of observations across different tasks allows the policy to learn a shared representation. All of these enable our method to achieve high data efficiency.
Moreover, Fig.~\ref{fig_flow_ade_fde} shows that the in-box flow prediction error of R1+H0 is much higher than R1+H30, while in Table~\ref{tab_succ_few} the success rate difference between R1+H0 and R1+H30 is not significant. This suggests that our action policy is robust to imprecise flow, due to using predicted flow for training.

\subsection{Failure Modes Analyze}

% The ablation results on our action policy demonstrate the contribution of each component to the overall method. 
DP3 shows a high success rate on single tasks but struggles to distinguish between the bowl positions in Pick Bowl \#0-3 (Table~\ref{tab_succ_main}), resulting in occasionally moving to incorrect positions that correspond to another Pick Bowl task. This issue becomes more severe with fewer demonstrations (Table~\ref{tab_succ_few}).

RISE, which also uses the diffusion policy as DP3, does not experience similar failures in Pick Bowl \#0-3. 
However, both RISE and DP3 failed to show generalization in \#4-6 but consistently moved to the bowl's position as in training.
% to perform grasping. 
% Furthermore, RISE shows limited capability in discerning fine-grained spatial relationships. This is manifested in occasional safety violations caused by collisions with the table in the Pick Bowl and Fold Cloth tasks. The insufficient bowl lifting height exhibited by RISE in Pick Bowl \#0, which leads to failures when placing the bowl onto the platform, further demonstrates this limitation.
Furthermore, RISE occasionally triggers safety violations by crashing into the table in the Pick Bowl and Fold Cloth tasks.
RISE also tends to lift the bowl with insufficient height, causing failures in Pick Bowl \#0.
% , causing failures in placing the bowl onto the platform. 
We attribute these issues to the insufficient number of demonstrations for RISE, considering that the original work employed 50 demonstrations per task. 
% caused by moving towards the wrong position corresponding to another Pick Bowl task. 
% RISE struggles to learn actions from only 10 demonstrations, given that the original work used 50 demonstrations per task. 

SUGAR, using a Transformer decoder as the action head, shows good spatial generalization in Pick Bowl \#4,5 even without pretraining. 
% SUGAR shows a higher success rate in \#4 than \#5 because \#4 has a similar initial action direction as \#1, while that of \#5 is different from the others. 
The pretraining of SUGAR increases the general success rate, especially for the deformable object task Fold Cloth. 
% especially for the Fold Cloth task with a deformable object.
% especially for discriminating irregular shapes of the deformable object in the Fold Cloth task. 
However, SUGAR consistently struggles in the Open Drawer task, which requires a detailed point cloud perception to produce accurate actions. 
Table~\ref{tab_succ_retry} shows the first-stage success rate of the Open Drawer task, where RISE and SUGAR have a lower retry success rate than Ours and DP3, which have point-level perception.
% , which have fine-grained point cloud perception. 
% This can be attributed to that Ours and DP3 have more fine-grained point cloud perception. 

Ours (w/o PC) consistently fails to hook the drawer handle in the Open Drawer task and fails put the trousers down in the Fold Cloth task.  
% Although it could successfully grasp the trousers in the Fold Cloth task, failures consistently occurred when putting them down. 
Ours (w/o PC) also occasionally triggers safety violations by colliding with the table. These issues highlight the importance of incorporating point cloud observations to adjust actions, rather than relying solely on the flow. 
Ours (w/o PF\&MP) has point cloud observation as a condition. Therefore, it shows a high success rate in the precision-demanding tasks: Open Drawer and Fold Cloth. However, we note a drop in the success rate in Pick Bowl \#4-6 caused by moving to the bowl positions as in training. This training-task overfitting is also observed in DP3 and RISE. Nonetheless, our method exhibits stronger generalization as a result of following the general motion provided by the flow.

\begin{figure}[t]
  \centering
  \begin{subfigure}{\linewidth}
    \centering
    \includegraphics[width=0.84\linewidth]{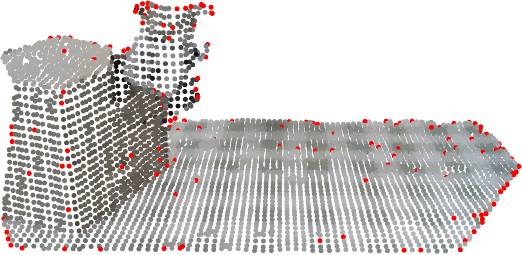}
    \caption{DP3 (Full Scene Point Cloud)}
    \label{fig_pc_key_dp3}
  \end{subfigure}
  
  \vspace{1em}
  
  \begin{subfigure}{\linewidth}
    \centering
    \includegraphics[width=0.84\linewidth]{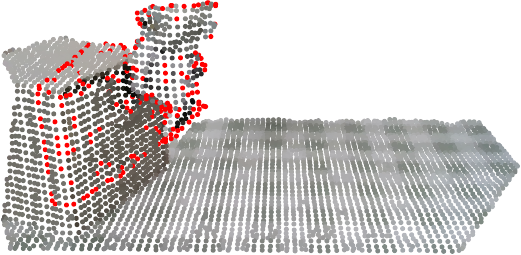}
    \caption{Ours (Local Cropped Point Cloud)}
    \label{fig_pc_key_ours}
  \end{subfigure}
  
  \caption{Max-pooling referenced points (red) in Open Drawer. }
  \label{fig_pc_key}
\end{figure}

\section{Discussion}
In this section, we discuss the underlying mechanisms of our method and how they address the issues in the prior work through the following research questions. 

% how our flow and local point cloud conditioned policy achieves generalization without compromising the performance for the precision-demanding tasks by addressing three research questions. 

\textbf{RQ2: Through what mechanism does cropping the point cloud enhance the policy performance in precision tasks?}
Compared to the DP3 point cloud encoder that provides point-level perception, RISE and SUGAR's group-level method is coarser-grained. Although this improves their overall scene understanding, they remain incapable of discerning small changes, which prevents them from completing tasks that require more than just coarse-level actions, such as Open Drawer. DP3 encoder uses max-pooling to extract features from the referencing points, providing precise point-level descriptions of the scene. However, selecting referencing points from the point cloud of the entire scene results in sparse, redundant, or even uninformative points. 
In Pick Bowl tasks, sometimes there are even no referencing points on the bowl. 
Fig.~\ref{fig_pc_key} compares the referencing points chosen from the whole scene and a local cropped point cloud. 
This explains why the original work of DP3 crops the table point cloud. 
The cropped point cloud has fewer points, resulting in more concentrated referencing points. Although the DP3 encoder still tends to select edge points as reference points in the cropped point cloud, the internal reference points become more densely distributed. The more reference points on the target object, the better its position and shape can be represented, especially for a noisy point cloud. 
% In return, the task success rate increases. 

% \textbf{RQ2: Is flow an appropriate condition to enhance the generalization ability of diffusion policy?}
\textbf{RQ3: How does conditioning on flow enhance policy generalization?}
Diffusion policy tends to overfit the training tasks \citep{he2025demystifying, wu2025afforddp}. Therefore, DP3 and RISE always move to a position associated with the training tasks and fail to generalize to unseen bowl positions in Pick Bowl \#4-6. 
The DP3 encoder with sparse referencing points is insensitive to variations in the scene, producing similar embeddings despite changes in the position of the bowl. This aggravates the overfitting of the diffusion policy. As a result, DP3 occasionally moves to a wrong position that is associated with another training task in Pick Bowl \#0-3, while RISE does not. Although our action policy also employs the diffusion policy, enabled by the flow, our method achieves both spatial and instance generalization. Our action policy without point cloud observation (w/o PC) shows generalization in Pick Bowl \#4-6 following the flow condition. These findings suggest that, rather than treating the flow as a dense label-like condition, the action is calculated based on the flow to follow the same motion, thereby achieving the generalization. 
% Therefore, flow is a better condition than the point cloud observation, which reduces overfitting and enhances generalization of the diffusion policy.

% % \textbf{RQ3: How effective is randomly masking the point cloud (MP) in alleviating overfitting of the diffusion policy?}
% \textbf{RQ4: In what way does balancing the reliance alleviate overfitting of the diffusion policy?}
% % Our action policy without point cloud observation (w/o PC) tends to hit the table when picking the bowl and performs poorly in Fold Cloth and Open Drawer. 
% % Our action policy with point cloud observation, but not setting the point cloud observation to 0 with a certain probability and not trained using the predicted flow (w/o PF\&MP), 
% Ours (w/o PF\&MP) tends to move towards the wrong position corresponding to another training task in the Pick Bowl tasks, while Ours and Ours (w/o PC) do not. These results indicate that the primary cause of diffusion policy overfitting is the excessive reliance on point cloud observations. They highlight the necessity of setting the point cloud to 0 (MP), thereby balancing the reliance between the point cloud and the flow to reduce overfitting. This is further validated by the ablation results in Table~\ref{tab_succ_few}. These results also indicate that training the action policy with the predicted flow (PF) does not introduce notable overfitting to the predicted flow. 
% % However, for tasks that require accurate actions, such as Open Drawer, this augmentation method could lead to an accuracy drop. Therefore we suggest 

\textbf{RQ4: How effective is balancing the reliance in alleviating overfitting of the diffusion policy?}
% The ablated variant, which is not trained with predicted flow and randomly masked point cloud (w/o PF\&MP), is the vanilla version of the flow and the cropped point cloud conditioned action policy. 
% Compared to the ablated version of our action policy that is trained with PF but does not have point cloud observation (Ours w/o PC), the version 
The ablated version of our flow-condition policy with point cloud observation but without PF and MP (Ours w/o PF\&MP) tends to move towards an incorrect bowl position corresponding to the training Pick Bowl tasks. While the ablated version with PF but without point cloud observation (Ours w/o PC) and the full version of our action policy that uses MP do not have such an overfitting problem. 
These results indicate that (1) training the policy with predicted flow (PF) does not introduce notable overfitting, (2) the primary cause of the overfitting is the reliance on point cloud observations.
Therefore, it is necessary to undermine the point cloud by random masking (MP), thereby balancing the reliance between the point cloud and the flow to reduce overfitting. 
% This is further validated by the ablation results in Table~\ref{tab_succ_few}. 
% Therefore, it is necessary to randomly mask the point cloud (MP) to let the policy know that the point cloud observation is not always reliable, thereby balancing the reliance between the point cloud and the flow to reduce overfitting. 
% The full version of our policy with MP significantly alleviates the seen-position overfitting problem, especially with limited demonstrations (Figure~\ref{tab_succ_few}). 
% These results highlight the necessity of randomly masking the point cloud, thereby balancing the reliance between the point cloud and the flow to reduce overfitting. 

In summary, for the policy network, it is essential to (1) leverage flow to guide the motion for generalization, (2) apply centering and cropping to the point cloud to enable fine-grained action adjustment based on the concentrated observation, and (3) randomly mask out the point cloud to achieve a balanced conditioning and reduced overfitting. 
% on the flow and point cloud.

\section{Conclusion}
We propose a cross-embodiment scene flow prediction model SFCr, which could generalize to unseen embodiments with segmentation and achieve high cross-embodiment data efficiency without segmentation. 
% The robot/hand segmentation is not indispensable when both robot and human data are available, but it is vital when robot data is missing. 
% follows the general flow motion and adjusts the action based on local point cloud observation 
We propose a flow and cropped point cloud conditioned action policy, FCrP, that achieves high data efficiency and both spatial and instance generalization. 
The overall system SFCrP reduces the IL data requirement, making it possible to generalize to scenarios only seen in human videos. 
% Our flow alignment method enables the action policy to predict multiple times conditioning on the same flow, achieving a flexible and robust execution. 
% Our method shows a higher success rate than the baselines across different task types, including scenarios only seen in human videos. 
We empirically analyze group-level and point-level perception of point clouds, as well as the source of diffusion policy overfitting. 
% We thoroughly ablated our method to illustrate the contribution of each component to the overall performance. 
We thoroughly ablated our method to illustrate how our method addresses the issues in the prior work, providing valuable insights for future research.
% However, our work still has limitations. 
However, we did not consider the flow-length variations caused by demonstrations having different speeds and motions.
% variations in human and robot demonstrations. 
% We set the range of point cloud cropping as a hyperparameter. 
% We set a constant cropping range across our experiment, but the best cropping range may vary per task. 
% Furthermore, even with a cropped point cloud, the DP3 encoder may still reference edge points. 
We train the flow prediction model SFCr using random query points, but execute it with in-box queries. The resulting difference in query point distribution causes an increase in the flow prediction error. Moreover, we mask the point cloud to balance the condition, while this method could reduce the policy performance on precision tasks.
These problems remain open challenges for future research. 
% We do not assume the human and robot demonstrations have the same speed, nor do we consider aligning the flow to account for the varying lengths caused by different speeds. Future work involves aligning the flow with different lengths.
% We train the flow prediction model using random query points, but execute it with in-box queries. The resulting gap in prediction error remains to be addressed in future work.
% Moreover, applying rotation to keep the robot gripper upright after centering and cropping the point cloud to enhance the generalization ability further is another promising direction of future work. 
\section*{ACKNOWLEDGMENT}
We would like to thank the Robotics@ANUComputing group for providing the experimental equipment and workspace. We also thank Hanna Kurniawati, Miaomiao Liu, and Rahul Shome for their helpful feedback and valuable discussions. 

% \twocolumn[\newpage]
\small
\bibliographystyle{unsrtnat}
\bibliography{ref}

\end{document}